\documentclass{article}
\usepackage{spconf,amsmath,graphicx}
\usepackage{booktabs}
\usepackage{color}

\title{AUTOMATIC ESTIMATION OF ICE BOTTOM SURFACES FROM RADAR IMAGERY}
\name{Mingze Xu$^{\dagger}$ \qquad David J. Crandall$^{\dagger}$ \qquad Geoffrey C. Fox$^{\dagger}$ \qquad John D. Paden$^{\star}$}
\address{$^{\dagger}$ School of Informatics and Computing, Indiana University, Bloomington, IN USA \\
$^{\star}$ Center for Remote Sensing of Ice Sheets, University of Kansas, Lawrence, KS USA}
\begin{document}
%
\maketitle
\begin{abstract}
    Ground-penetrating radar on planes and satellites now makes it practical to
    collect 3D observations of the subsurface structure of the polar ice sheets,
    providing crucial data for understanding and tracking global climate change.
    But converting these noisy readings into useful observations is generally
    done by hand, which is impractical at a continental scale. In this paper,
    we propose a computer vision-based technique for extracting 3D ice-bottom
    surfaces by viewing the task as an inference problem on a probabilistic
    graphical model. We first generate a seed surface subject to a set of
    constraints, and then incorporate additional sources of evidence to refine
    it via discrete energy minimization. We evaluate the performance of the
    tracking algorithm on 7 topographic sequences (each with over 3000 radar
    images) collected from the Canadian Arctic Archipelago with respect to
    human-labeled ground truth.
\end{abstract}


\vspace{-2pt}
\section{Introduction}

\begin{figure*}
    \begin{center}
        \includegraphics[width=17.5cm]{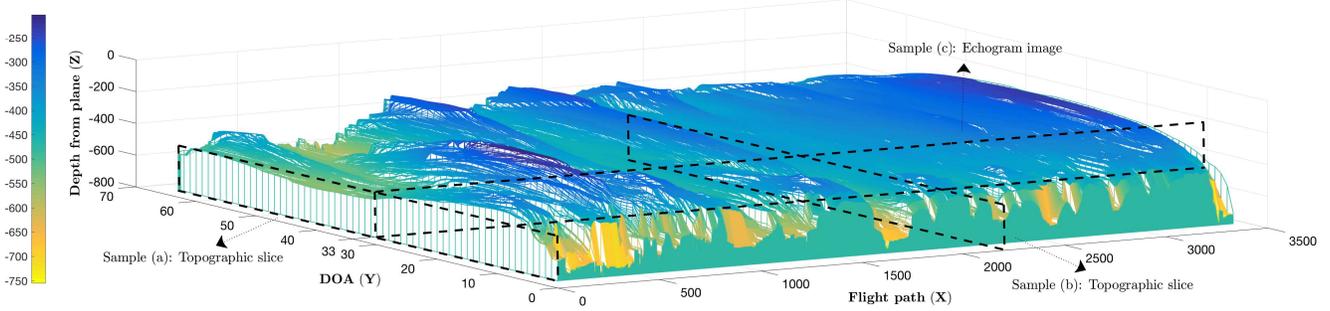}
    \end{center}
    \vspace{-15pt}
    \caption{Illustration of our task. Radar flies along the X-axis, 
    collecting noisy evidence about the ice surface distance and depth immediately below it. This yields a 2D echogram (Sample (c)), with depth on one axis and flight
    path on the other,
    and prior work has used these echograms to estimate 2D ice structure but only along the flight path.  Our approach also includes (very noisy) evidence
    from either side of the radar, yielding a sequence of 2D topographic slices (e.g.\ Sample (a) and (b)). 
    Each slice  is represented in polar coordinates,
    where Y- and Z-axis denote the direction of arrival of radar waves and
    the distance from each voxel to plane, respectively. We combine this noisy evidence with prior information to produce
    3D ice reconstructions.}
    \label{fig:overall}
    \vspace{-10pt}
\end{figure*}

Scientists increasingly use visual observations of the world in their work:
astronomers collect telescope images at unprecedented scale~\cite{szalay2001world},
biologists image live cells~\cite{jaiswal2003long, stephens2003light},
sociologists record social interactions~\cite{wedekind2000cooperation},
ecologists collect large-scale remote sensing data~\cite{bamber2013new}, etc.
Although progress in technology has made \textit{collecting} this imagery
affordable, actually \textit{analyzing} it is often done by hand. But with
recent progress in computer vision, automated techniques may soon work well
enough to remove this bottleneck, letting scientists analyze visual data more
thoroughly, quickly, and economically.

As a particular example, glaciologists need large-scale data about the polar
ice sheets and how they are changing over time in order to understand and
predict the effects of melting glaciers.
Aerial ground-penetrating radar systems have been developed that can
fly over an ice sheet and collect evidence about its subsurface structure.
The raw radar return data is typically mapped into 2D radar echogram images
which are easier for people to interpret, and then manually labeled for
important semantic properties (ice thickness and structure,
bedrock topography, etc.) in a slow, labor-intensive
process~\cite{freeman2010automated, ilisei2012technique, ferro2013automatic,
mitchell2013semi}. Some recent work has shown promising results on the specific problem of layer-finding in 2D echograms~\cite{crandall2012layer,
lee2014estimating, carrer2017automatic}, although the accuracy is still far
below that of a trained human annotator. The echograms are usually quite noisy
and complex, requiring experience and intuition that is difficult to
encode in an algorithm. Using echograms as input data also inherently limits
the analysis to the ice structure immediately under the radar's 
flight path.

In this paper we take an alternative approach, using additional data collected
by the radar in order to actually estimate the 3D structure of the ice sheet,
including a large area on either side, instead of simply tracing 2D
cross-sections (Figure~\ref{fig:overall}). In particular, the Multichannel Coherent Radar Depth Sounder
(MCoRDS) instrument~\cite{rodriguez2014advanced} uses three transmit beams
(left, nadir, right) to collect data from below the airplane
and to either side (for a total swath width of about 3km). Although an expert
may be able to use intuition and experience to produce a
reasonable estimate of the 3D terrain from this data, the amount of weak
evidence that must be considered at once is overwhelming. As with
structure-from-motion in images~\cite{disco2013pami}, this gives
automatic algorithms an advantage: while humans are better
at using intuition to estimate from weak evidence, algorithms can
consider a large, heterogeneous set of evidence to make better overall decisions.

We formulate the problem as one of discrete energy minimization in order to
combine weak evidence into a 3D reconstruction of the bottom of the ice sheet.
We first estimate layer boundaries to generate a seed surface, and then
incorporate additional sources of evidence, such as ice masks, surface digital
elevation models, and optional feedback from humans to refine it.
We investigate the performance of the algorithm using
ground truth from humans, showing that our technique significantly
outperforms several strong baselines.


\vspace{-5pt}
\section{Related work}

Detecting boundaries between material layers in noisy radar images is important
for glaciology. Semi-automated and automated methods have been introduced for
identifying features of subsurface imaging. For example, in echograms from Mars,
Freeman et al.~\cite{freeman2010automated} find layer boundaries by applying
band-pass filters and thresholds to find linear subsurface structures, while
Ferro and Bruzzone \cite{ferro2011novel} identify subterranean features using
iterative region-growing. For the specific case of ice, Crandall et al.~\cite{crandall2012layer}
detect the ice-air and ice-bottom layers in echograms along the flight path by
combining a pretrained template model for the vertical profile of each layer
and a smoothness prior in a probabilistic graphical model. Lee et al.~\cite{lee2014estimating}
present a more accurate technique that uses Gibbs sampling from a joint
distribution over all possible layers. Carrer and Bruzzone~\cite{carrer2017automatic}
reduce  computational complexity with a divide-and-conquer strategy. In contrast
to the above work which all infers 2D cross-sections, we attempt to reconstruct
3D subsurface features and are not aware of other work that does this. We pose
this as an inference problem on a Markov Random Field similar to that proposed
for vision problems (e.g.\ stereo~\cite{felzenszwalb2006efficient}), except
that we have a large set of images and wish to produce a 3D surface, whereas
they perform inference on a single 2D image at a time.

\vspace{-5pt}
\section{Methodology}

As the radar system flies over ice, it collects a
sequence of topographic slices $I = \{I_1, \cdots, I_l\}$ that characterizes
the returned radar signals (Figure 1). Each slice $I_i$ is a 2D radar image that describes a
distribution of scattered energy in
polar coordinates (with dimensions $\phi \times \rho)$ at a discrete position $i$ of the aircraft along its flight path.
Given such a topographic sequence of dimension $l \times \phi \times \rho$,
we wish to infer the 3D ice-bottom surface.
We parameterize the surface as a sequence of slices $S = \{S_1, \cdots, S_l\}$
and
$S_i = \{s_{i,1}, \cdots, s_{i,\phi}\}$, where $s_{i,j}$ denotes the 
row coordinate of the boundary of the ice-bottom for column $j$ of slice $i$,
and $s_{i,j} \in [1,\rho]$
since the ice-bottom layer can occur anywhere within a column.

\vspace{-4pt}
\subsection{A graphical model for surface extraction}

Because radar  is so noisy, our goal is to find a surface that
not only fits the observed data well but that is also smooth
and satisfies other prior knowledge.  We formulate this as an
inference problem on a Markov Random Field.  In particular, we
look for a surface that minimizes an energy function,
\begin{align} \label{eq:1}
        E(S|I) = & \sum\limits_{i=1}^l \sum\limits_{j=1}^\phi  \psi_1(s_{i,j}|I) \ + \\ 
& \sum\limits_{i=1}^l \sum\limits_{j=1}^\phi \sum_{i' \in \pm 1} \sum_{j' \in \pm 1} \psi_2(s_{i,j}, s_{i+i', j+j'})  
\end{align}
where
$\psi_1(\cdot)$
defines a unary cost function which measures how well a given labeling agrees
with the observed image in $I$, and $\psi_2(\cdot, \cdot)$ defines a
pairwise interaction potential on the labeling which encourages the surface to
be continuous and smooth. Note that each column of each slice contributes one
term to the unary part of the energy function, while the pairwise terms 
are a summation over the four neighbors of a column (two columns on either side
within the same slice, and two slices within the same column in neighboring slices).

\vspace{8pt}
\noindent
\textbf{Unary term.}
Our unary term $\psi_1(\cdot)$ consists of three parts,
\begin{equation} \label{eq:5}
    \psi_1(\cdot) = \psi^{temp}(\cdot) + \psi^{air}(\cdot) + \psi^{bin}(\cdot).
\end{equation}
First, similar to~\cite{crandall2012layer}, we define a template model $T$
of fixed size $1 \times t$ (we use $t = 11$ pixels) for the vertical profile
of the ice-bottom surface in each slice. For each pixel $p$ in the template,
we estimate a mean $\mu_p$ and a variance $\sigma_p$ on greyscale intensity
assuming that the template is centered at the location of the ice-bottom
surface, suggesting a template energy,
\begin{equation} \label{eq:2}
    \psi^{temp}(s_{i,j}|I) = \sum\limits_{p \in T} (I(s_{i,j} + p) - \mu_p)^2 / \sigma_p.
\end{equation}
We learn the parameters of this model with a small set of labeled
training data.


\newcommand{\ra}[1]{\renewcommand{\arraystretch}{#1}}
\begin{table}\centering
{\footnotesize{
    \ra{1.1}
    \begin{tabular}{@{}lcccccc@{}} \toprule
        & & \multicolumn{2}{c}{Error} & \phantom{a}& \multicolumn{2}{c}{Precision}\\
        \cmidrule{3-4} \cmidrule{6-7}
        & & \textbf{Mean} & \textbf{Median Mean} && \textbf{1 pixel} & \textbf{5 pixels} \\ \midrule
        \multicolumn{7}{l}{(a) Ice-bottom surfaces:} \\
        Crandall \cite{crandall2012layer} & & 101.6 & 95.9 & & 0.2\% & 2.5\% \\
        Lee \cite{lee2014estimating} & & 35.6 & 30.5 & & 3.6\% & 29.9\% \\
        Ours with \textbf{DV} & & 13.3 & 13.4 & & 20.2\% & 58.3\% \\
        Ours with \textbf{TRW} & & 11.9 & 12.2 & & 35.9\% & 63.9\% \\ \midrule
        \multicolumn{7}{l}{(b) Bedrock layers:} \\
        Crandall \cite{crandall2012layer} & & 75.3 & 42.6 & & 0.5\% & 21.5\% \\
        Lee \cite{lee2014estimating} & & 47.6 & 36.6 & & 2.2\% & 20.5\% \\
        Ours with \textbf{TRW} & & 4.1 & 4.2 & & 28.8\% & 81.4\% \\
        \bottomrule
    \end{tabular}
}}
    \vspace*{-5pt}
    \caption{Error in terms of the mean and median mean absolute
    column-wise difference compared to ground truth, in pixels. Precision is
    the percentage of correct labeled pixels.}
    \vspace{-10pt}
\end{table}


\begin{figure}[t]
    \begin{minipage}[b]{1.0\linewidth}
        \centering
        \centerline{\includegraphics[height=2.5cm, width=8.5cm]{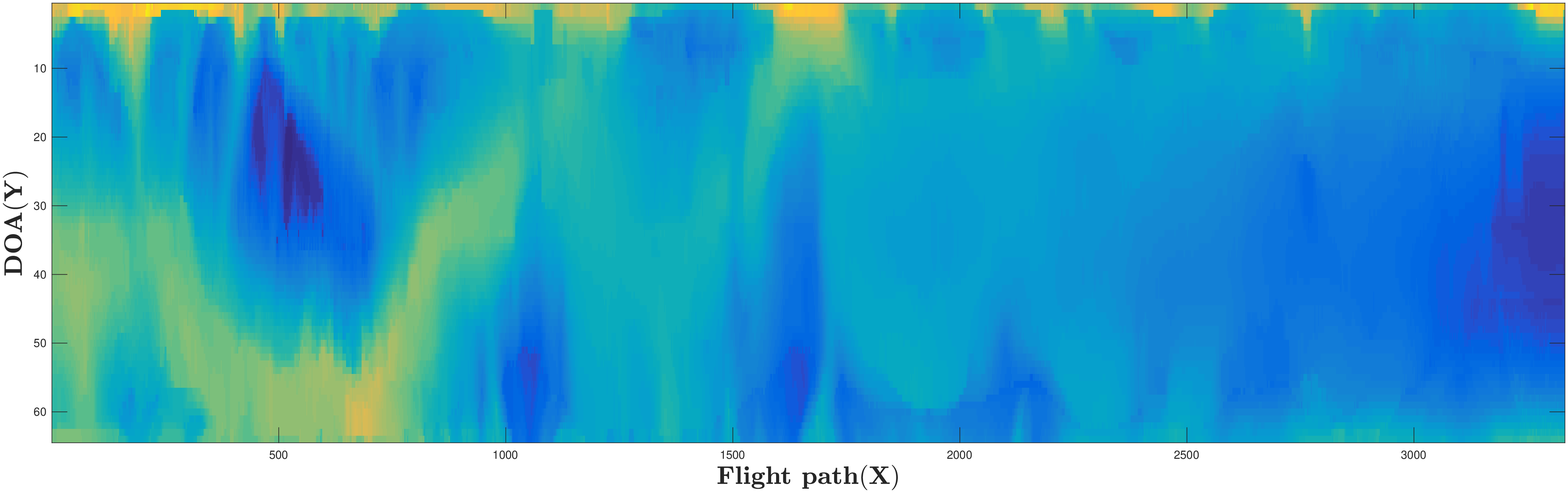}}
        \vspace{-25pt}
        \centerline{\textcolor{white}{Ground truth}}\medskip
        \vspace{8pt}
    \end{minipage}
    \begin{minipage}[b]{1.0\linewidth}
        \centering
        \centerline{\includegraphics[height=2.5cm, width=8.5cm]{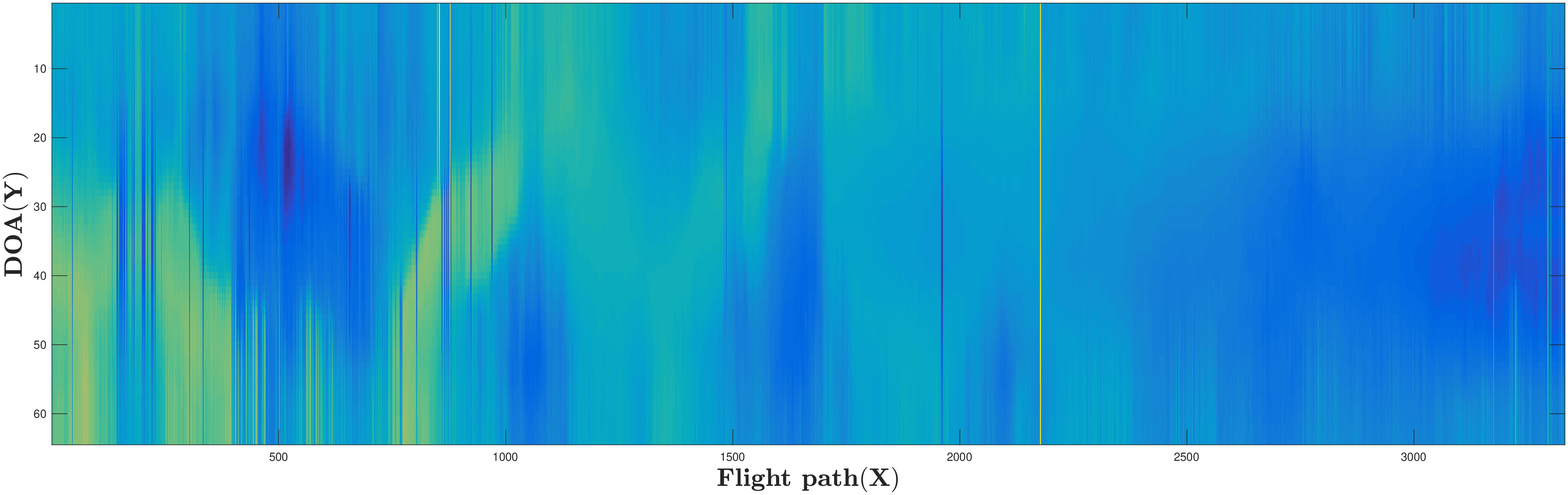}}
        \vspace{-25pt}
        \centerline{\textcolor{white}{Result of \cite{lee2014estimating}}}\medskip
        \vspace{7pt}
    \end{minipage}
    \begin{minipage}[b]{1.0\linewidth}
        \centering
        \centerline{\includegraphics[height=2.5cm, width=8.5cm]{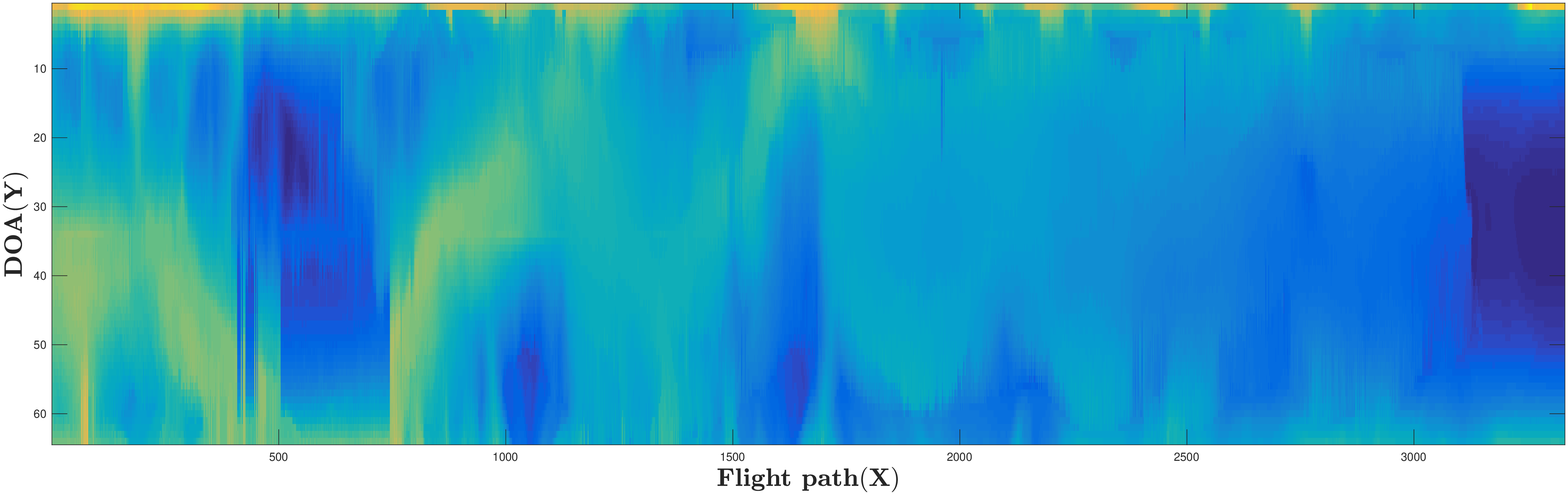}}
        \vspace{-25pt}
        \centerline{\textcolor{white}{Ours with DV}}\medskip
        \vspace{8pt}
    \end{minipage}
    \begin{minipage}[b]{1.0\linewidth}
        \centering
        \centerline{\includegraphics[height=2.5cm, width=8.5cm]{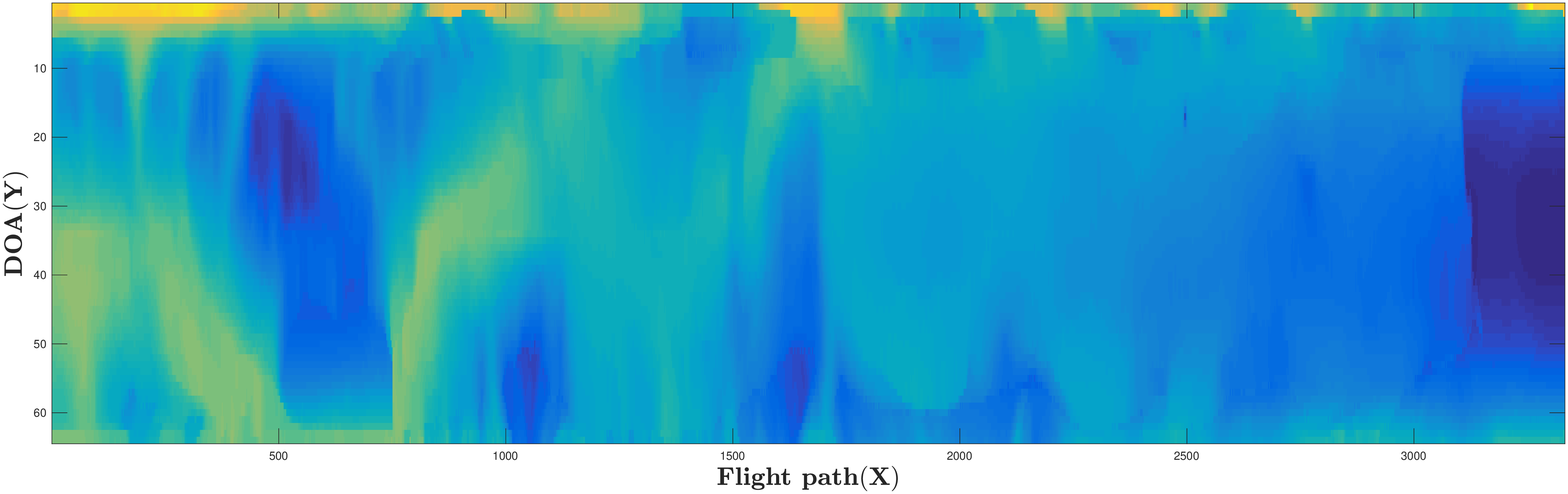}}
        \vspace{-25pt}
        \centerline{\textcolor{white}{Ours with TRW}}\medskip
        \vspace{8pt}
    \end{minipage}
    \vspace*{-18pt}
    \caption{Results of the ice-bottom surface extracting on a sample dataset. The color
    represents the depth from plane (Z).}
    \label{fig:surface}
    \vspace{-10pt}
\end{figure}

Second, to capture the fact that the ice-bottom surface should always be below the
ice-air surface by a non-trivial margin, we add a cost  to penalize intersecting
surfaces,
\begin{equation} \label{eq:3}
    \psi^{air}(s_{ij}) =
    \begin{cases}
        \, \quad \quad +\infty & s_{i,j} - a_{i,j} < 0 \\
        \, \, \, \, \, \, \quad \quad 0 & s_{i,j} - a_{i,j} > \tau \\
        \, \tau - |s_{i,j} - a_{i,j}|  & \mbox{otherwise,}
    \end{cases}
\end{equation}
with $a_{i,j}$ the label of the air-ice boundary of slice
$i$, column $j$.

Finally, we incorporate an additional weak source of evidence produced
by the radar system. The \emph{bottom bin} gives a constraint on a 
\textit{single} column in each slice, specifying a single coordinate $(j, b_i)$
that the true surface boundary must be below. Despite how weak this evidence
is, it helps to distinguish between the ice-air and ice-bottom surface boundary
in practice.
Formally, we formulate
this cost function as,
\begin{equation} \label{eq:4}
    \psi^{bin}(s_{i,j}) =
    \begin{cases}
        \, +\infty & s_{i,j} < b_i \\
        \, \, \, \, \, \, 0 & \mbox{otherwise.}
    \end{cases}
\end{equation}


\vspace{6pt}
\noindent
\textbf{Pairwise term.}
The ice-bottom surface is
encouraged to be smooth 
across both adjacent columns and adjacent slices,
\begin{equation} \label{eq:6}
    \psi_2(s, \hat{s}) =
    \begin{cases}
        -\beta_j \ln \mathcal{N} (s - \hat{s}; \, 0, \hat{\sigma})  & |s - \hat{s}| < \alpha \\
        \quad \quad \quad \quad +\infty & \mbox{otherwise,}
    \end{cases}
\end{equation}
where $\hat{s}$ denotes the labeling of an adjacent pixel of ($i, j$), and
parameters $\alpha$ and $\hat{\sigma}$ are learned from labeled training data.
Parameter $\beta_j$ models smoothness on
a per-slice basis, which is helpful if some slices are known to be noisier
than others (or set to a constant if this information is not known).
This term models the similarity of the labeling of two adjacent pixels by a
zero-mean Gaussian that is truncated to zero outside a fixed interval $\alpha$.
Since all parameters in the energy function are considered penalties, we
transform the Gaussian probability to a quadratic function by using a negative
logarithm.

\vspace{8pt}
Our energy function introduces several important improvements over
that of Crandall et al.~\cite{crandall2012layer} and Lee et
al.~\cite{lee2014estimating}.  First, while their model gives all
pairs of adjacent pixels the same pairwise weight ($\beta$), we have
observed that layers in different slices usually have particular
shapes, such as straight lines and parabolas, depending on the local ice
topography. By using a dynamic weight $\beta_j$, we can roughly
control the shape of the layer and adjust how smooth two adjacent
pixels should be. More importantly, those techniques consider a single
image at a time, which could cause discontinuities in the ice reconstruction.
We correct this by defining pairwise terms along both the intra- and inter-slice
dimensions.

\vspace{-6pt}
\subsection{Statistical inference}

The minimization of equation ($\ref{eq:1}$) can 
be formulated as discrete energy minimization on a first-order
Markov Random Field (MRF)~\cite{koller2009probabilistic}. 
Given the large size of this MRF, we use  Sequential Tree-reweighted
Message Passing (TRW)~\cite{kolmogorov2006convergent}, which breaks the MRF
into several monotonic chains, and perform belief propagation (BP) on
each chain. TRW only passes messages within each of these chains, rather than
to all four directions (like Loopy BP~\cite{murphy1999loopy}). Benefiting from
this, TRW converges faster and requires half as much memory as traditional
message passing methods. We assign a row-major order for pixels in the graph
and define the monotonic chains based on this order. In each iteration, TRW first
passes messages in increasing order, and then back in
decreasing order. 
We pre-define a maximum number
of iterations to be the same as the width of each slice, $\phi$, which allows
evidence from one side of the slice to reach the other.
When message
passing is finished, we assign a label to each pixel in row-major order:
for pixel $(i, j)$, we choose the label $s_{i,j}$ that minimizes
$\boldsymbol{M}(s_{i,j}) + \psi_1(s_{i,j}) + \psi_2(s_{i,j}, s_{i,j-1}) + \psi_2(s_{i,j}, s_{i-1,j})$,
where $\boldsymbol{M}(s_{i,j})$ is the summation of messages from four adjacent neighbors.


The usual implementation of TRW has time complexity $O(l \phi \rho^2)$ for
each loop. To speed this up, we use linear-time generalized distance
transforms~\cite{felzenszwalb2006efficient},
yielding a total running time of $O(l \phi \rho L)$ where
$L$ is the number of iterations.
This is possible because of our 
pairwise potentials are log-Gaussian. 

\vspace{-15pt}
\section{Experiments}

\begin{figure}[t]
    \begin{minipage}[b]{1.0\linewidth}
        \centering
        \centerline{\includegraphics[height=2.0cm, width=8.5cm]{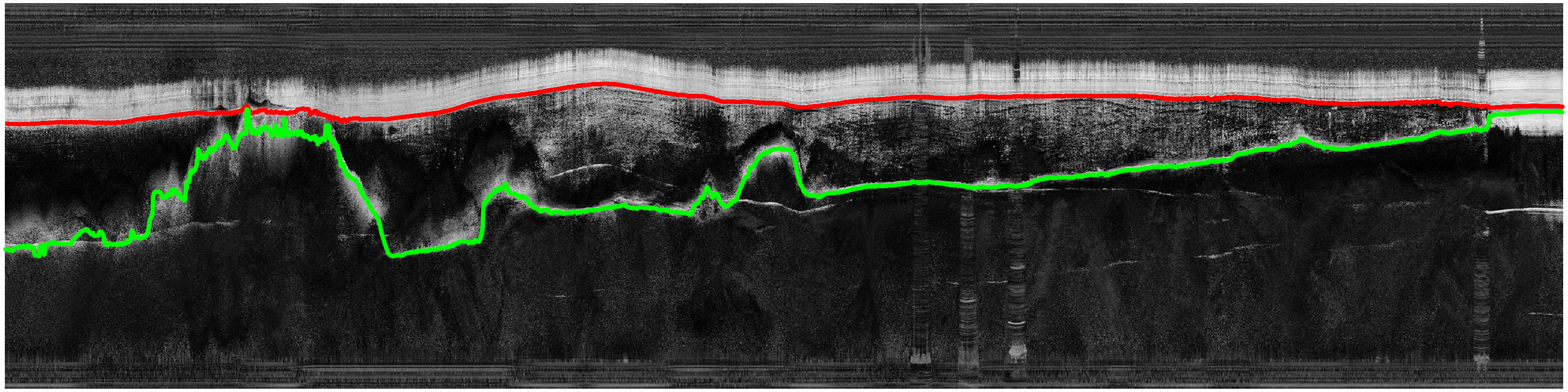}}
        \vspace{-20pt}
        \centerline{\textcolor{white}{Ground truth}}\medskip
        \vspace{1.5pt}
    \end{minipage}
    \begin{minipage}[b]{1.0\linewidth}
        \centering
        \centerline{\includegraphics[height=2.0cm, width=8.5cm]{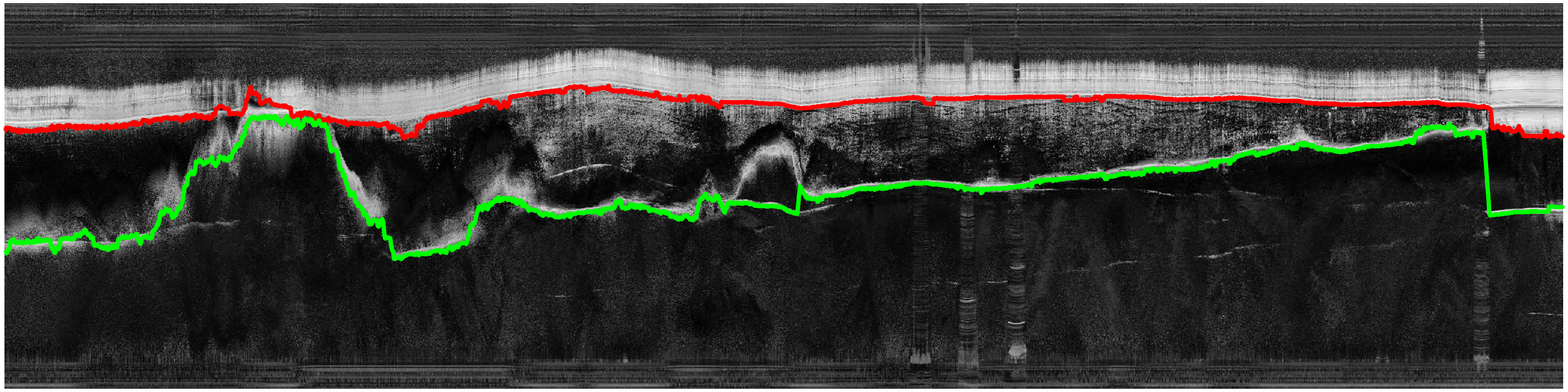}}
        \vspace{-20pt}
        \centerline{\textcolor{white}{Result of \cite{crandall2012layer}}}\medskip
    \end{minipage}
    \begin{minipage}[b]{1.0\linewidth}
        \centering
        \centerline{\includegraphics[height=2.0cm, width=8.5cm]{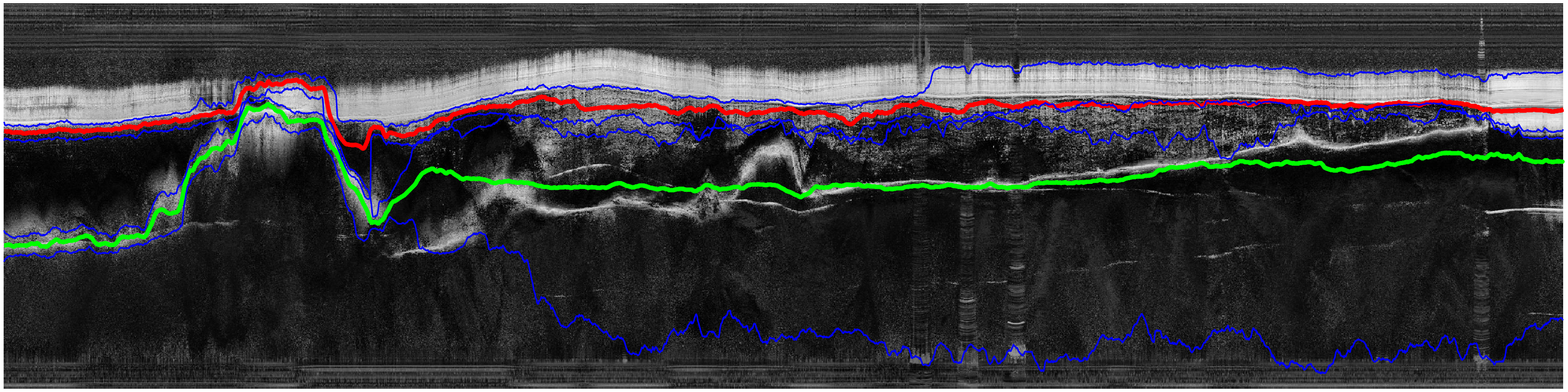}}
        \vspace{-20pt}
        \centerline{\textcolor{white}{Result of \cite{lee2014estimating}}}\medskip
    \end{minipage}
    \begin{minipage}[b]{1.0\linewidth}
        \centering
        \centerline{\includegraphics[height=2.0cm, width=8.5cm]{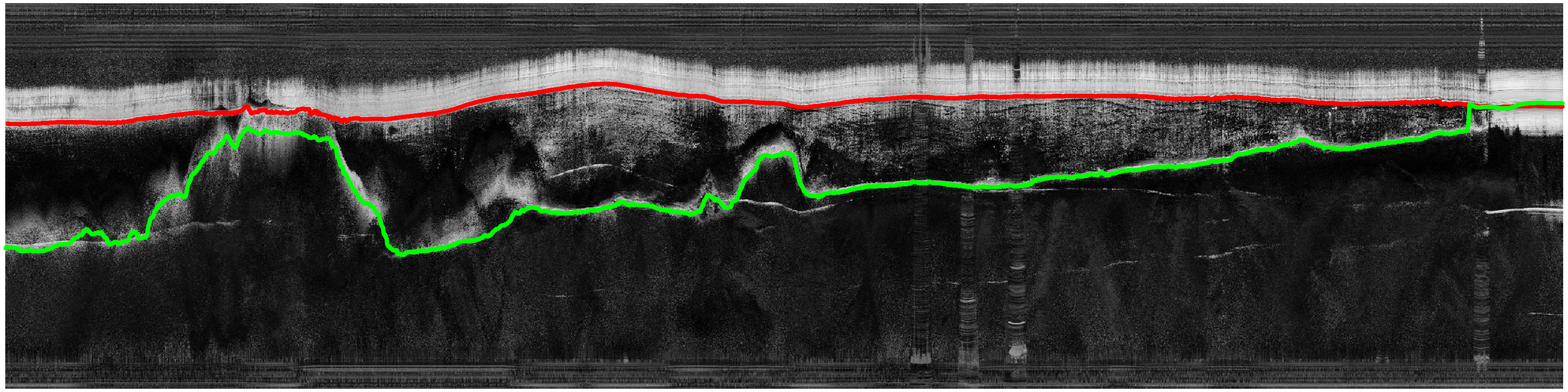}}
        \vspace{-20pt}
        \centerline{\textcolor{white}{Ours with TRW}}\medskip
    \end{minipage}
    \vspace*{-15pt}
    \caption{Results of the bedrock layer finding on a sample echogram. In each image,
    the upper (red) boundary is the ice-air layer, and the lower
    (green) boundary is the ice-bottom layer. The ice-air layer in our
    result is from the radar.}
    \label{fig:echogram}
    \vspace{-10pt}
\end{figure}

We tested our surface extraction algorithm on the basal topography of
the Canadian Arctic Archipelago (CAA) ice caps, collected by the
Multichannel Coherent Radar Depth Sounder (MCoRDS)
instrument~\cite{rodriguez2014advanced}.  We used a total of 7
topographic sequences, each with over 3000 radar images which
corresponds to about 50km of flight data per sequence. For these
images, we also have the associated ice-air surface ground truth, a subset
(excluded from the testing data)
of which we used to learn the parameters of the template model and the weights
of the binary costs.

We then ran inference on each topographic sequence and measured the accuracy
by comparing our estimated surfaces to the ground truth, which was produced
by human annotators.
However, these labels are not always accurate at the pixel-level,
since the radar images are often full of noise, and some boundaries
simply cannot be tracked precisely even by experts.
To relax the effect of inaccuracies in
ground truth, we consider a label to be correct when it is within 
a few pixels. 
We evaluated with three summary statistics: mean
deviation, median mean deviation, and the percentage of correct labeled pixels
over the whole surface (Table 1(a)). The mean error is about 11.9
pixels and the median-of-means error is about 12.2 pixels. The
percentage of correct pixels is 35.9\%, or about 63.9\% within 5 pixels,
which we consider the more meaningful statistic
given noise in the ground truth.

To give some context, we compare our results with three baselines.
Since no existing methods solve the 3D reconstruction problem that we consider here,
we adapted three methods from 2D layer finding to the 3D case.
%
Crandall et al.~\cite{crandall2012layer} use a fixed weight for the
pairwise conditional probabilities in the Viterbi algorithm, which
cannot automatically adjust the shape of the layer in each image
slice.  Lee et al.~\cite{lee2014estimating} generate better results by
using Markov-Chain Monte Carlo (MCMC).  However, neither of these approaches
considers constraints between adjacent slices. We introduce
Dynamic Viterbi (DV) as an additional baseline that incorporates a
dynamic weight for the pairwise term, but it still lacks the ability
to smooth the whole surface in 3D.  As shown in Table 1(a) and Figure
2, our technique performs significantly better than any of these
baselines on 3D surface reconstruction.  We also used our technique to
estimate layers in 2D echograms, so that we could compare directly to
the published source code of~\cite{crandall2012layer,
  lee2014estimating} (i.e.\ using our approach to solve the problem
they were designed for). Figure 3 and Table 1(b) present results,
showing a significant improvement over these baselines also.

Similar to~\cite{crandall2012layer, lee2014estimating}, additional evidence
can be easily added into our energy function. For instance, ground truth
data (e.g.\ ice masks) may be available for some particular slices, and human
operators can also provide feedback by marking true surface boundaries for
a set of pixels. Either of these can be implemented by putting additional
terms into the unary term defined in equation ($\ref{eq:5}$).
%

\vspace{-5pt}
\section{Conclusion}
To the best of our knowledge, this paper is the first to propose an automated
approach to reconstruct 3D ice features using graphical models.
We showed our technique can
 effectively estimate ice-bottom surfaces from noisy
radar observations.
This technique also demonstrated its accuracy and efficiency in producing
bedrock layers on radar echograms against the state-of-the-art.

\vspace{-8pt}
\section{Acknowledgements}
This work was supported in part by the National Science Foundation (DIBBs
1443054, CAREER IIS-1253549), and used the Romeo cluster, supported by Indiana
University and NSF RaPyDLI 1439007. We acknowledge the use of data from CReSIS
with support from the University of Kansas and Operation IceBridge (NNX16AH54G).

\bibliographystyle{IEEEbib}
\bibliography{strings,refs}

\end{document}